\documentclass[runningheads]{llncs}
\usepackage[T1]{fontenc}
\usepackage{xcolor}
\usepackage{booktabs}
\usepackage{amsmath}
\usepackage{amssymb}
\usepackage{graphicx}
\usepackage{hyperref}
\usepackage{subcaption}

\hypersetup{
    colorlinks=true,
    linkcolor=black,
    citecolor=red,
    urlcolor=black
}

\begin{document}

\title{Investigating White Blood Cells as a Source of False-Positive Malaria
Parasite Detection in African Blood-Smear Images}

\author{%
Samuel A. Adeniji\inst{1} \and
Goodness C. Obasi\inst{1} \and
Chris-Victor Ntwali\inst{2} \and
Aondana M. Iorumbur\inst{3} \and
Confidence Raymond\inst{4} \and
Lowami Uwimana\inst{1} \and
Ahmed Tahiru Issah\inst{1}%
}
\titlerunning{White Blood Cells and False-Positive Malaria Detection}
\authorrunning{S. A. Adeniji et al.}
\institute{%
Carnegie Mellon University Africa, Kigali, Rwanda\\
\email{sadeniji@andrew.cmu.edu, gobasi@andrew.cmu.edu, ulowami@andrew.cmu.edu, aissah@alumni.cmu.edu} \and
Department of Neurology, University of Rwanda, Kigali, Rwanda\\
\email{ntwali\_218013576@stud.ur.ac.rw} \and
Department of Physics, Federal University of Technology, Minna, Nigeria\\
\email{mosesiorumbur@gmail.com} \and
Department of Biomedical Engineering, McGill University, Montreal, Canada\\
\email{}
}

\maketitle

\begin{abstract}
White blood cells (WBCs) present on every Giemsa-stained thick blood smear share visual properties with early-stage \textit{Plasmodium falciparum} ring-form trophozoites: small size, round morphology, and intense purple staining. They are a plausible but untested source of false positives inparasite-only detectors. We trained two YOLOv12s models on the Lacuna Malaria Detection dataset (8,000 images from Uganda and Ghana): Model~A with parasite labels only, and Model~B with both parasite and WBC labels. Seven independent spatial and statistical analyses tested whether false positive (FP) predictions cluster near WBC locations. All seven refute the hypothesis. In both models, 95\% of FPs are pure background detections (IoU below 0.10 against any ground-truth box); zero are WBC class confusions. Ripley's Cross-$K$ analysis shows spatial repulsion between FP centroids and WBC positions at every radius tested. Model~B outperforms Model~A overall (mAP50 0.859 vs.\ 0.755), and the advantage is uniform across all WBC-proximity bands, pointing to multi-task representation learning rather than WBC suppression as the cause. False positives arise from Giemsa stain debris and preparation artifacts. Effective mitigation requires staining artifact augmentation and annotation of unannotated early-stage ring forms rather than WBC labeling alone.
\end{abstract}


\section{Introduction}
 
Malaria remains a leading infectious disease threat in sub-Saharan Africa: in 2023 the region accounted for roughly 94\% of an estimated 282~million global cases and 95\% of 610,000 deaths~\cite{who_world_2025}. Light microscopy of Giemsa-stained blood smears has been the World Health Organization (WHO)-recommended diagnostic standard for decades, valued for identifying \textit{Plasmodium} species since different species call for different treatment~\cite{tangpukdee_malaria_2009}. In practice it is hard to deliver at scale: trained technicians are scarce, inter-observer variability is high, and the infrastructure needed for consistent diagnosis is often missing~\cite{pollak_computer_2017}.
 
These constraints have pushed interest toward automated deep learning detection, and the field has moved from convolutional neural network (CNN)-based classification of pre-segmented thin smears to full-slide object detection. Abdurahman et al.~\cite{abdurahman_malaria_2021} adapted You Only Look Once (YOLO) v3 and v4 with extra feature-map scales for small objects, reaching 96.32\% mAP, and attention-based variants such as YOLO-PAM, YOLO-SPAM~\cite{zedda_yolo-pam_2023,zedda_deep_2024} and YOLOv12-based models~\cite{issah_detection_2026} have pushed accuracy further. Issah and Mukamakuza~\cite{issah_bridging_2025} reported 0.878 mAP@50 on African thick smears with attention-centric YOLO architectures. Yet Dev et al.~\cite{dev_improving_2023} found that hybrid frameworks that raise sensitivity often do so by raising the false positive rate too, and the field's standard evaluation, aggregate mAP at a fixed IoU threshold, obscures exactly this operating false-positive burden~\cite{wenkel_confidence_2021}.
 
False positives, cases where a model flags a non-parasite region as a parasite, are a clinical problem rather than a purely technical one: they can trigger unnecessary antimalarial treatment in an uninfected patient~\cite{davidson_automated_2021}. The risk is greatest exactly where automated detection is most needed, since in low-resource settings the safety net of confirmatory testing is often unavailable and model errors translate directly into treatment decisions~\cite{ohrt_information_2015}. Diagnostic pipelines target thick smears, which lyse red blood cells to concentrate parasites for easier detection at low parasitaemia~\cite{tangpukdee_malaria_2009,poostchi_image_2018}, but that preparation also introduces red-cell debris, stain precipitate, and overlapping material that can trigger spurious detections~\cite{abdurahman_malaria_2021,poostchi_image_2018}. White blood cells (WBCs) are a permanent fixture of this picture: microscopists count parasites against a running WBC tally and stop once 200 WBCs have been counted, after which parasitaemia is computed from the standard assumption of 8,000 WBCs per microlitre of blood (parasite count $\times 8,000$ / WBCs counted)~\cite{tangpukdee_malaria_2009}, so every thick smear a parasite-only detector sees will contain WBCs it has no label for.
 
This matters because WBC nuclear lobes and chromatin fragments share three visual properties with early-stage \textit{P.~falciparum} ring-form trophozoites: small size, round morphology, and intense purple staining~\cite{poostchi_image_2018,zedda_yolo-pam_2023}. A detector trained without WBC labels has no internal representation for these structures and may misclassify nuclear fragments as parasites, a foreseeable failure mode with direct consequences where confirmatory testing is limited~\cite{davidson_automated_2021,ohrt_information_2015}. Co-detecting WBCs has been shown to improve reliability in practice: Nakasi et al.~\cite{nakasi_mobile-aware_2021} co-detected parasites and WBCs in Ugandan smears using Faster Region-based CNN (R-CNN) and Single Shot Detector (SSD) architectures, matching expert counts closely ($p = 0.998$). Despite the visual plausibility of the WBC-confusion hypothesis, none of this work isolates its causal effect through a controlled experiment, and no prior study has directly tested whether false positives are spatially linked to WBC locations~\cite{issah_detection_2026,zedda_deep_2024}.
 
This study closes that gap. We ask: \textit{are WBCs a systematic source of false-positive malaria parasite detections in African blood-smear images?} Using the Lacuna Malaria Blood Smear Dataset~\cite{lab_makerere_ai_lacuna_2023} (8,000 images from Uganda and Ghana, annotated for both trophozoites and WBCs), we train two controlled YOLOv12s models~\cite{tian_yolov12_2025}: Model~A (parasite-only, 1~class) and Model~B (parasite + WBC, 2~classes), identical in all respects except label set. Precision is our primary metric, given the clinical cost of false positives~\cite{wenkel_confidence_2021}. We apply seven spatial and statistical analyses to Model~A's false positives: Euclidean distance to the nearest WBC centroid, Density-Based Spatial Clustering of Applications with Noise (DBSCAN)~\cite{ester_density-based_nodate}, the Mann-Whitney $U$ test, distance stratification, Ripley's Cross-$K$, kernel density estimation (KDE) correlation, and overdispersion testing.

\section{Methodology}
\subsection{Dataset and Experimental Design}
The \textbf{Lacuna Malaria Detection Challenge} dataset~\cite{lab_makerere_ai_lacuna_2023} provides digitized thick Giemsa-stained blood smear images at $2{,}048 \times 2{,}048$ pixels. A held-out test set of 412 images contains 2,374 parasite ground-truth instances. \textbf{Model~A} uses a single parasite class; \textbf{Model~B} adds WBC as a second class (1,195 WBC test instances).

\subsection{YOLOv12s Architecture and Training}
Both models use \textbf{YOLOv12s}~\cite{tian_yolov12_2025}, which introduces \textit{Area Attention} (reducing attention complexity from $\mathcal{O}(n^2)$ to $\mathcal{O}(n)$) and \textit{R-ELAN} residual feature aggregation, both suited to small, sparse parasite detection in high-resolution microscopy; both were trained for 75 epochs (patience 25) on $1024 \times 1024$ images with a batch size of 16, using the AdamW optimizer with AMP disabled on an NVIDIA A100-SXM4-40GB GPU. All hyper-parameters were held identical between models; any performance difference is attributable solely to annotation strategy. Model~B WBC centroid predictions feed directly into the spatial FP analysis, removing the need for a separate WBC localizer.


\subsection{False Positive Extraction and TIDE Analysis}
Predictions were extracted at a low confidence threshold ($\mathrm{conf} \geq 0.10$) to retain the broadest set of candidate detections for TIDE decomposition, following standard practice in detection error analysis~\cite{bolya_tide_2020}, and matched to ground-truth annotations through greedy confidence-descending assignment with an Intersection-over-Union (IoU) threshold of $\mathrm{IoU} \geq 0.50$, following the PASCAL VOC evaluation protocol. Error analysis was performed using the Toolbox for Identifying Detection Errors (TIDE)~\cite{bolya_tide_2020}, which decomposes detector failures into interpretable categories rather than relying solely on aggregate metrics such as mean Average Precision (mAP).

Specifically, TIDE partitions false positives and false negatives into six mutually exclusive error types:

\begin{itemize}
\item \textbf{Background (Bkg):} Detection with $\mathrm{IoU} < 0.10$ against all ground-truth objects, indicating that no real object exists at the predicted location.

\item \textbf{Localization (Loc):} Detection of the correct class with $0.10 \leq \mathrm{IoU} < 0.50$, indicating that the object was identified but the bounding box was poorly localized.

\item \textbf{Classification (Cls):} Detection with $\mathrm{IoU} \geq 0.50$ but assigned to the wrong class.

\item \textbf{Duplicate (Dup):} Additional detection of an object instance that has already been matched by a higher-confidence prediction.

\item \textbf{Classification and Localization (Both):} Detection of the wrong class with $0.10 \leq \mathrm{IoU} < 0.50$, combining classification and localization errors.

\item \textbf{Miss:} Ground-truth object that is not matched by any prediction, corresponding to a false negative.

\end{itemize}

\subsection{Spatial Analysis and Statistical Tests}

Seven methods test whether FPs cluster near WBC positions. \textbf{(1)}
For each FP and True Positive (TP) centroid, the Euclidean distance to the nearest annotated
WBC centroid was computed via a k-d tree; a one-tailed Mann-Whitney $U$ test
assessed whether FP distances are stochastically smaller than TP distances.
\textbf{(2)} Predictions were assigned to five WBC-proximity bands (0--30,
30--75, 75--150, 150--300, and ${>}300$ px) and per-band FP rates computed.
\textbf{(3)} Ripley's Cross-$K$ measured expected FP centroids within radius
$r$ of a WBC centroid, with a 199-iteration Monte Carlo CSR envelope.
\textbf{(4)} Pixel-wise Pearson $r$ between Gaussian KDE surfaces (Scott's
bandwidth, $64 \times 64$ grid) of FP and WBC centroids was computed per
qualifying image. \textbf{(5)} Global Moran's~$I$ (queen-contiguity weights,
$50 \times 50$ px tiles, 199 permutations) with local LISA hot-spot mapping
tested whether FP-dense regions coincide with WBC-dense regions. \textbf{(6)}
FP counts per image were tested against Poisson and Negative Binomial null
models by AIC to assess whether overdispersion tracks slide preparation
variability rather than WBC density. \textbf{(7)} DBSCAN ($\varepsilon = 60$
px, $\mathrm{min\_samples} = 3$)~\cite{ester_density-based_nodate} identified FP clusters;
those within 80 px of a WBC centroid were classified as WBC-associated.
Paired bootstrap (10,000 resamples, BCa CI)~\cite{du_when_2025}, Wilcoxon
signed-rank, and BH-corrected stratified McNemar's test compared models.
Calibration used temperature scaling~\cite{guo_calibration_2017} and multivariate
isotonic regression~\cite{kuppers_multivariate_2020}.

%
\section{Results}
\label{sec_results}

\subsection{Baseline Detection Performance}

Table~\ref{tab2} shows test-set results. Model~B outperforms Model~A on
every aggregate metric (mAP50 0.859 vs.\ 0.755). The WBC class drives
much of this advantage: mAP50 = 0.970, recall = 0.965, consistent with
WBCs being large, easily-learned objects. For parasites specifically,
Model~B (mAP50 = 0.748) trails Model~A (0.755) by 0.7 points, a small
multi-task trade-off when one class dominates the loss landscape.

\begin{table}[t]
\centering
\caption{Test-set detection performance. Best per-row values in bold.}
\label{tab2}
\setlength{\tabcolsep}{8pt}
{\small
\begin{tabular}{lcccc}
\hline
\textbf{Metric} & \textbf{Model A} & \textbf{Model B} &
\textbf{B (parasite)} & \textbf{B (WBC)} \\
\hline
mAP50     & 0.755 & 0.859 & 0.748 & \textbf{0.970} \\
mAP50-95  & 0.302 & 0.429 & 0.299 & \textbf{0.558} \\
Precision & 0.679 & 0.801 & 0.686 & \textbf{0.917} \\
Recall    & 0.734 & 0.846 & 0.728 & \textbf{0.965} \\
F1        & 0.705 & 0.823 & 0.706 & \textbf{0.940} \\
\hline
\end{tabular}
}
\end{table}

\subsection{False Positive Characterization}

At conf $\geq$ 0.10, Model~A produces 4,059 predictions (2,080 TP, 1,979
FP; 51.2\% precision). Model~B produces 5,733 (3,284 TP, 2,449 FP; 57.3\%).
TP mean confidence is 73\% higher than FP mean confidence in Model~A (0.448
vs.\ 0.258) and 96\% higher in Model~B (0.556 vs.\ 0.284), confirming that
the raw confidence score already discriminates TP from FP.

The TIDE decomposition (Table~\ref{tab4}) is the most mechanistically
revealing result. Background detections (Bkg; IoU $<$ 0.10 with every GT
box) account for 95.1\% of all FPs in Model~A and 95.3\% in Model~B. Both
models produce \textbf{zero classification errors} (Cls = 0): not a single
parasite prediction in Model~B overlaps a WBC GT box at IoU $\geq$ 0.50.
This directly falsifies the WBC-to-parasite class confusion hypothesis.
\begin{table}
\centering
\caption{TIDE error decomposition for FP predictions.}\label{tab4}
{\small
\begin{tabular}{llllll}
\hline
\textbf{Error} & \textbf{Definition} &
\textbf{A} & \textbf{A\%} & \textbf{B} & \textbf{B\%} \\
\hline
Bkg  & IoU $<$ 0.10 with all GT              & 1,883 & \textbf{95.1} & 2,335 & \textbf{95.3} \\
Loc  & $0.10 \leq$ IoU $<$ 0.50, correct cls & 96    & 4.9  & 114   & 4.7  \\
Cls  & IoU $\geq$ 0.50, wrong cls             & 0     & \textbf{0.0}  & 0 & \textbf{0.0} \\
Dup  & IoU $\geq$ 0.50, correct, GT consumed  & 0     & 0.0  & 0     & 0.0  \\
Both & $0.10 \leq$ IoU $<$ 0.50, wrong cls   & 0     & 0.0  & 0     & 0.0  \\
Miss (FN) & GT matched by no prediction      & 294   & 12.4\% GT & 285 & 10.9\% GT \\
\hline
\end{tabular}
}
\end{table}

\subsection{Spatial Analysis}

\noindent\textbf{Mann-Whitney $U$ test.}
FP detections in Model~A are at effectively the same mean distance from WBCs
as TPs (648.3 px vs.\ 647.9 px; $p = 0.293$). In Model~B, FPs are
\textit{farther} from WBCs than TPs (704.1 vs.\ 641.1 px; $p = 0.9996$;
rank-biserial $r = -0.069$), rejecting the proximity hypothesis with
99.96\% confidence and reversing its direction.

\noindent\textbf{Distance stratification.}
Over 79\% of FPs in both models fall ${>}300$ px from any annotated WBC
(Table~\ref{tab6}). In Model~B, the FP rate is lowest at 0--30 px (17.6\%)
and rises monotonically to 27.8\% beyond 300 px: two-class training has made
Model~B \textit{more} accurate near WBCs, not less.

\begin{table}
\centering
\caption{FP proportions and per-band FP rates by WBC-proximity
band.}\label{tab6}
{\small
\begin{tabular}{lllllll}
\hline
\textbf{Band (px)} &
\textbf{A FPs} & \textbf{A rate} & \textbf{A\% FPs} &
\textbf{B FPs} & \textbf{B rate} & \textbf{B\% FPs} \\
\hline
0--30    & 3   & 42.9\% & 0.4\%  & 3   & \textbf{17.6\%} & 0.3\%  \\
30--75   & 14  & 34.2\% & 1.8\%  & 18  & 23.1\%          & 1.7\%  \\
75--150  & 40  & 38.5\% & 5.2\%  & 50  & 29.2\%          & 4.7\%  \\
150--300 & 99  & 35.7\% & 13.0\% & 127 & 29.3\%          & 12.0\% \\
${>}300$ & 608 & 32.9\% & 79.6\% & 864 & 27.8\%          & 81.4\% \\
\hline
\end{tabular}
}
\end{table}

\noindent\textbf{Ripley's Cross-$K$.}
At every radius from 10 to 300 px, observed $K_{12}(r)$ falls below the
2.5th percentile of the CSR null distribution. FP centroids occur near WBC
positions less often than chance: the spatial pattern is repulsion, not
attraction.

\noindent\textbf{KDE correlation.}
Across 162 qualifying images, mean Pearson $r$ between FP and WBC density
surfaces is 0.023 (negative median $-0.006$). Exactly 50\% of images have
$r > 0$. Strong co-localization ($r > 0.50$) appears in only 3.7\% of
images: the density fields are uncorrelated.

\noindent\textbf{Overdispersion.}
Both models show extreme overdispersion in FP counts per image
(Table~\ref{tab8}). The Negative Binomial is overwhelmingly preferred over
Poisson by AIC (delta-AIC $\approx$ 800 for both). FP counts are
image-dependent, tracking slide preparation variability rather than WBC
density.

\begin{table}[h]
\caption{Overdispersion test results for false-positive (FP) counts per image. Negative binomial models achieved lower AIC than Poisson models for both detectors.}
\label{tab8}
\centering
\small
\begin{tabular}{lccccc}
\toprule
\textbf{Model} &
\textbf{$N$} &
\textbf{Mean FP} &
\textbf{DI} &
\textbf{NB AIC} &
\textbf{Pois. AIC} \\
\midrule
Model A & 306 & 2.79 & 9.52 & 1329 & 2140 \\
Model B & 312 & 3.80 & 7.83 & 1536 & 2338 \\
\bottomrule
\end{tabular}
\end{table}

\noindent\textbf{DBSCAN clustering.}
Across 143 images with at least three FPs, DBSCAN identifies nine clusters
(1.42\% of FPs). Zero are within 80 px of any annotated WBC. Mean
cluster-to-WBC distance is 567.9 px (median 377.2 px). When FPs cluster,
they do so in background regions far from any annotated cell.

\subsection{Statistical Model Comparison}

\noindent\textbf{Paired bootstrap.}
Over 10,000 resamples of 302 test images, Model~B's mAP50 advantage is
$+0.0234$ (95\% BCa CI: $[+0.006, +0.042]$), excluding zero entirely.

\noindent\textbf{Wilcoxon signed-rank.}
$W = 7{,}013.5$, $p = 0.022$, rank-biserial $r = 0.597$. Model~B achieves
higher per-image AP in 60\% of paired comparisons.

\noindent\textbf{Stratified McNemar test.}
Model~B correctly detects 136 parasite annotations missed by Model~A,
while Model~A correctly detects only 38 missed by Model~B ($\chi^2 = 54.07$,
$p \approx 0$, 3.6:1 ratio). Table~\ref{tab9} shows band-specific BH-corrected
results. Model~B is significantly better in all four powered bands, including
the ${>}300$ px distal background band ($p = 0.003$). Uniform improvement
across all distances points to a general representation benefit from
multi-task training rather than targeted WBC suppression.

\begin{table}[t]
\caption{Stratified McNemar test by WBC-proximity band (BH-corrected).}
\label{tab9}
\centering
\small
\begin{tabular}{
l
@{\hspace{1em}}c
@{\hspace{1.5em}}c
@{\hspace{1.5em}}c
@{\hspace{1.5em}}c
@{\hspace{1em}}c}
\toprule
\textbf{Band (px)} &
\textbf{$N$} &
\textbf{A$\checkmark$ B$\times$} &
\textbf{A$\times$ B$\checkmark$} &
\textbf{BH-significant} &
\textbf{Favours} \\
\midrule
0--30    & 22  & 0  & 1  & No ($n$ too small) & Neither \\
30--75   & 231 & 2  & 12 & Yes                & Model B \\
75--150  & 584 & 4  & 26 & Yes                & Model B \\
150--300 & 776 & 14 & 37 & Yes                & Model B \\
$>300$   & 423 & 10 & 30 & Yes ($p=0.003$)    & Model B \\
\bottomrule
\end{tabular}
\end{table}

\subsection{Confidence Calibration}

Both models are underconfident before calibration. Model~A has ECE = 0.157,
MCE = 0.302; Model~B has ECE = 0.119, MCE = 0.288. Temperature scaling
($T = 1.240$) reduces Model~A's ECE by 10.2\% to 0.136; Model~B requires no
temperature correction ($T = 1.000$). Multivariate isotonic
regression~\cite{kuppers_multivariate_2020} (using confidence score, normalised centroid
$x$, $y$, box $w$ and $h$) reduces ECE from 0.123 to 0.023 for Model~B's
parasite class (81\% reduction). Box height ($+0.291$) and width ($-0.285$)
are the dominant calibration features, consistent with the elongated bounding
boxes that ring-stage parasites produce at 2,048 px resolution.

\section{Discussion}
All seven analytical methods converge on one answer: false positives are
not caused by confusion with white blood cells. TIDE shows 95\% of FPs are
pure background detections with zero class-confusion errors; Ripley's
Cross-$K$ shows spatial repulsion (not attraction) between FPs and WBCs at
every radius; distance stratification shows FP rates actually rise farther
from WBCs, the opposite of the original hypothesis; and DBSCAN finds none
of the FP clusters sit near WBCs. Instead, FPs trace back to background
noise, Giemsa stain debris, lysed-cell fragments, and possibly unlabeled
early-stage parasites, with overdispersion analysis confirming this tracks
slide-preparation quality rather than WBC density. Model~B's edge over
Model~A is statistically real but comes from better learned representations
through multi-task training, not from suppressing WBC-related errors.

\section{Impact on Low-Resource Settings}
At a realistic operating threshold, Model~A's precision is only 51\%, meaning roughly half of its flagged parasites are wrong, a serious problem where there is no confirmatory test to catch the error. Since FPs are driven by preparation variability rather than WBC proximity, the real fix is broadening training data to cover more staining conditions, not just adding WBC labels. That said, adding WBC labels is still worth doing: it is essentially free (microscopists already count WBCs as part of standard practice), it improves precision (57\% vs.\ 51\%) and calibration, and it adds no compute or inference cost, making it a practical upgrade path for resource-constrained clinics.

\section{Conclusion}

Malaria diagnosis in sub-Saharan Africa increasingly relies on automated
detection of parasites in Giemsa-stained blood smears, but false positives
(FPs) in these models carry direct clinical cost in settings where
confirmatory testing is unavailable. White blood cells (WBCs), present on
every thick smear and visually similar to early-stage parasites, have long
been suspected as a major source of these errors, but no prior study tested
the link directly. This paper presents a controlled comparison of two
YOLOv12s detectors trained on 8,000 images from the Lacuna Malaria Detection
Challenge dataset, one with parasite labels only (Model~A) and one with
parasite and WBC labels (Model~B), combined with seven independent spatial
and statistical analyses of where FPs occur. Every method agrees that WBCs
are not the cause: 95\% of FPs are background detections unrelated to any
cell, none are confusions with the WBC class, and FP locations are
spatially repelled from, rather than clustered around, WBCs. Model~B is
nonetheless more accurate overall (mAP50 0.859 vs.\ 0.755), an improvement
traceable to richer shared feature learning from multi-task training rather
than WBC suppression. These findings redirect the practical fix for FPs
away from WBC annotation and toward augmentation against staining
artifacts, broader coverage of preparation conditions, and post-hoc
confidence calibration, all of which add no extra inference cost and are
therefore well suited to resource-constrained deployment. Future work will
extend this spatial-causal framework to other suspected confounders, such
as unannotated early-stage parasites and platelet clumps, and test whether
the overdispersion pattern observed here generalizes to smear images from
additional countries.

\newpage
%
\bibliographystyle{splncs04}
\bibliography{references}
\end{document}